\documentclass[journal]{IEEEtai}

\usepackage[colorlinks,urlcolor=blue,linkcolor=blue,citecolor=blue]{hyperref}

\usepackage{color,array}

\usepackage{amsmath,amssymb}
\usepackage{graphicx}
\usepackage{subcaption}
\usepackage{comment}

\newcommand{\D}{\mathcal{D}}
\newcommand{\Dx}{\mathcal{D}_{|x}}
\newcommand{\Dy}{\mathcal{D}_{|y}}
\newcommand{\rev}[1]{\textcolor{black}{#1}}

\begin{document}

\title{On the Role of Priors in Bayesian Causal Learning} 

\author{Bernhard C. Geiger\thanks{Bernhard C. Geiger (geiger@ieee.org) is with the Signal Processing and Speech Communication Laboratory, Graz University of Technology, Inffeldgasse 16c, 8010 Graz, Austria and with the Know Center Research GmbH, Sandgasse 34, 8010 Graz, Austria.}, \emph{Senior Member, IEEE} and Roman Kern\thanks{Roman Kern is with the Institute for Interactive Systems and Data Science, Graz University of Technology, Sandgasse 36, 8010 Graz, Austria and with the Know Center Research GmbH, Sandgasse 34, 8010 Graz, Austria.}}


\maketitle

\begin{abstract}
In this work, we investigate causal learning of independent causal mechanisms from a Bayesian perspective. Confirming previous claims from the literature, we show in a didactically accessible manner that unlabeled data (i.e., \rev{cause realizations}) do not improve the estimation of the parameters defining the mechanism. Furthermore, we observe the importance of choosing an appropriate prior for the cause and mechanism parameters, respectively. Specifically, we show that a factorized prior results in a factorized posterior, which resonates with Janzing and Schölkopf's definition of independent causal mechanisms via the Kolmogorov complexity of the involved distributions and with the concept of parameter independence of Heckerman et al.
\end{abstract}

\begin{IEEEImpStatement}
Learning the effect from a given cause is an important problem in many engineering disciplines, specifically in the field of surrogate modeling, which aims to reduce the computational cost of numerical simulations. Causal learning, however, cannot make use of unlabeled data \rev{-- i.e., cause realizations --}  if the mechanism that produces the effect is independent from the cause. In this work, we recover this well-known fact from a Bayesian perspective. Our work further suggests that the prior distribution of cause and mechanism parameters should factorize, since such a distribution may be most efficient for learning, especially in the small-data regime.
\end{IEEEImpStatement}


\section{Introduction}
Causality has seen an increase in interest in the AI community, as it allows to address issues such as robustness and fairness in machine learning~\cite{scholkopf2022causality}.
A key property of causation is its asymmetric nature, which for example can be exploited for causal discovery. The causal direction also has important implications on what can be learned from data~\cite{Schölkopf2012}.

Causal learning problems, i.e., learning the effect from a cause, or learning the mechanism that transforms a cause into an effect, are manifold in science and engineering. In mechanical engineering, for example, applying a force (cause) to a metallic object leads to deformation, resulting in changed geometric dimensions or residual stress (effect). In material science, the structure and composition (cause) of a crystal determine its properties, such as conductivity or energy (effect). In these examples, deformation and structure-property relationships (mechanisms) are usually represented by first principles models, the simulation of which is often computationally costly. Therefore, substantial efforts are devoted to training surrogate models that can replace these simulations. These surrogate models require \emph{causal learning}, since they are used to predict the effect from the cause. \rev{Other examples for causal learning exist in natural language processing, cf.~\cite{Jin_2021} and} automatic speech recognition: The audio signal available to the automatic speech recognition system (cause) should be used to predict the transcript (effect), modelling human hearing (mechanism), cf.~\cite{Gabler_CausalityASR}.

Learning in the causal direction suffers from a big caveat, however: In a semi-supervised setting\footnote{\rev{Semi-supervised learning means that parameters are inferred from a dataset that contains both labeled and unlabeled instances. We consider an instance \emph{labeled} if it contains the value of the cause $x$ \emph{and} the value of the effect $y$. If only the cause values are recorded, we call the instance \emph{unlabeled}.}}, realizations of the cause $x$ do not help learning the mechanism $x\to y$ if it is independent from the cause, cf.~\cite[Sec.~2.1.2]{Schölkopf2012}. Indeed, the authors of~\cite{Janzing2015} investigated learning a bijective, monotonic mapping between cause and effect and, using results from information geometry, showed that \rev{realizations of $x$} can only help in the anti-causal setting~\cite[Th.~4]{Janzing2015}\rev{, i.e., when they are effect realizations}. In causal learning, \rev{cause realizations} can only help learning the mechanism $x\to y$ if, in addition to cause realizations $x$, also unlabeled effect realizations $z_y$, produced by a different mechanism $y\to z_y$, are given~\cite{vonKügelgen20a,vonKügelgen19}. \rev{Even generative models, which learn the joint distribution of causes and effects, are claimed to be less effective for causal learning than for anti-causal learning~\cite{Blöbaum_2015}.}

All these results hinge on the assumption that the mechanism $x\to y$ is independent of the cause $x$. The authors of~\cite{Janzing2015} declared independence if the cause and the slope (or logarithmic slope) of the function are uncorrelated, while the authors of~\cite{Janzing2010} defined an independent causal mechanism (ICM) as one whose algorithmic description cannot be compressed by knowing the algorithmic description of the cause. In terms of Kolmogorov complexity $K(\cdot)$, the joint distribution $\pi(x,y)$ of cause and effect then satisfies
\begin{equation}\label{eq:Kolmogorov}
    K(\pi(x,y)) \stackrel{+}{=} K(\pi(x)) + K(\pi(y|x))
\end{equation}
where $\stackrel{+}{=}$ implies that the equality holds up to a constant that may depend on the choice of the Turing machine, cf.~\cite[eq.~(4)]{vonKügelgen20a}.

In this work, we investigate causal learning of an ICM from a Bayesian perspective (Section~\ref{sec:setup}). Specifically, we assume that both cause and mechanism are parameterized, and that we perform Bayesian inference to learn these parameters. Using both factorized and general priors for these parameters, we show in a didactically accessible way that \rev{cause realizations} do not help in learning the parameter of the mechanism (Section~\ref{sec:causal}) and may even slow down learning (Section~\ref{sec:experiments}). We furthermore show that a factorized prior distribution on the parameters results in a factorized posterior (Section~\ref{sec:factorized}), agreeing with the characterization of ICMs via Kolmogorov complexity (Section~\ref{sec:discussion}).

\section{Related Work}

The work closest to ours is~\cite{Wu2022}. In this paper, the authors investigated domain adaptation and semi-supervised learning in the causal and anti-causal direction, investigating in which settings cause realizations (of the target domain) are useful and at which rates the excess risk decreases. Similarly to our work, the authors start with a prior distribution over cause and mechanism parameters (see Section~\ref{sec:setup}). The authors of~\cite{Wu2022} then consider a two-step learning problem, where in the first step they learn the cause and mechanism parameters from available data, and then apply the learned parameters for predicting the effect from the cause (potentially on a target domain with shifted distributions). In contrast, in this work we consider only the first of these two steps and only the semi-supervised learning setting (i.e., we do not consider distribution shifts). However, while in~\cite[p.~18, center]{Wu2022} cause realizations are simply not considered in the posterior of the mechanism parameter, the focus of our Section~\ref{sec:causal} is to justify this step in a didactic manner for ICMs. Furthermore, while~\cite{Wu2022} does not specify the joint prior on the cause and mechanism parameters, we show in Sections~\ref{sec:factorized} and~\ref{sec:discussion} that a factorized prior agrees better with the assumption of an ICM. Our work thus addresses~\cite[Remark~10]{Wu2022}, acknowledging that prior selection is important especially in the small-data regime.

\rev{At the first glance, one of our main results -- that a factorized prior on the parameters results in a factorized posterior -- is reminiscent of the corresponding \emph{parameter independence} result in~\cite[eqs.~(18)-(20)]{Heckermann_1995}. Specifically, the authors showed that a factorized prior for the distribution parameters of discrete variables in a Bayesian network results in a factorized posterior if complete datasets are observed. In cases of missing data, this posterior independence does not hold in general, as they illustrate at the hand of an uninformative, factorized Dirichlet prior~\cite[Sec. 5.6]{Heckermann_1995}. We believe that this results from the fact that~\cite{Heckermann_1995} compares various candidate structures of the Bayesian network and, at no point, relies on the ICM assumption.}

Therefore, while~\cite{Wu2022} is more general than our work in the sense of considering domain adaptation in addition to semi-supervised learning and more technical in quantifying learning rates, our work justifies fundamental steps required by~\cite{Wu2022} and provides a novel perspective on prior selection in Bayesian causal learning. \rev{Compared~\cite{Heckermann_1995}, our work considers also incomplete data (i.e., cause realizations without effect realizations), and shows that posterior parameter independence holds under the ICM assumption.} Finally, our work is more general (but less technical) than~\cite{Janzing2015}, which investigates only deterministic mechanisms and has quite restrictive conditions for the mechanism to be considered independent.

\section{Setup and Notation}
\label{sec:setup}
We make the common abuse of notation and do not distinguish between random variables (RVs) and their realizations. We let $\pi(\cdot)$ denote probability densities given ``by nature'', and $p(\cdot)$ probability densities obtained from modelling. We do not distinguish between densities w.r.t.\ the Lebesgue measure or w.r.t.\ the counting measure.

We suppose a \emph{structural causal model} in which a cause $x$ is fed into an ICM $x\to y$. Considering a semi-supervised learning setting, we assume to have access to a set $\D=\{(x_i,y_i)\}_{i=1}^N$ of paired cause and effect realizations. We abbreviate the collections of causes and effects in $\D$ as $\Dx=\{x_i\}$ and $\Dy=\{y_i\}$, respectively. In addition to this fully labeled dataset $\D$, we further have access to \rev{a dataset $\D_x$ of cause realizations}, i.e., $\D_x=\{x_i\}_{i=N+1}^{N+M}$.

We assume that the (distribution of the) cause and the (conditional distribution induced by the) ICM are parameterized by parameters $\theta$ and $\psi$, respectively. We do not assume that cause realizations are drawn independently or have identical distributions. \textbf{We do, however, assume that the ICM operates independently and identically on every cause at its input, and that $\D_x$ and $\D$ are drawn independently from each other.} Mathematically, the (joint) distributions of $\D$ and $\D_x$ are given as
\begin{subequations}\label{eq:pi}
\begin{align}
    \pi(\D,\D_x|\theta,\psi) &= \pi(\D|\theta,\psi)\pi(\D_x|\theta,\psi)\label{eq:pi_joint}\\
    \pi(\D|\theta,\psi) &=  \pi(\Dx|\theta) \prod_{i=1}^N \pi(y_i|x_i,\psi) \notag\\
    &\qquad= \pi(\Dx|\theta) \pi(\Dy|\Dx,\psi)\label{eq:pi_d}\\
     \pi(\D_x|\theta,\psi) &= \pi(\D_x|\theta)\label{eq:pi_dx}
\end{align}
\end{subequations}
where the conditioning on the parameters indicates that the distributions $\pi$ are parameterized by $\theta$ and $\psi$, respectively, and where~\eqref{eq:pi_dx} indicates that the distribution of $\D_x$ only depends on the parameters of the cause, as implied by the ICM.

We consider causal learning, i.e., we aim to infer the parameter $\psi$ of the ICM from data $\D$ and $\D_x$. To this end, we pursue a Bayesian approach. Specifically, we define a prior distribution $p(\theta,\psi)$ on the parameters and study the behavior of the posterior distribution $p(\theta,\psi|\D,\D_x)$, using~\eqref{eq:pi} as the likelihood. At this stage, we make no assumption on the prior $p(\theta,\psi)$ except that it is proper, i.e., continuous and positive on its support. 

There is consensus in the literature that cause realizations cannot improve our estimates of the ICM, i.e., $\D_x$ does not help in estimating $\psi$. The following example, \rev{where cause realizations change our belief about the mechanism parameter}, appears to be in contrast with this consensus and sets the motivation for the forthcoming analyses:

\vspace{1em}
\textbf{Example.} Suppose that the cause has a Gaussian distribution with mean $\mu$ and standard deviation $\sigma$, hence $\theta=(\mu,\sigma)$, and that the mechanism is a simple addition, i.e., $y=x+\psi$. Suppose that we have only access to \rev{cause realizations} $\D_x$, from which we can estimate the mean $\mu$ and standard deviation $\sigma$. Suppose further that our prior $p(\theta,\psi)$ has a large portion of the probability mass concentrated on the event $\psi=\mu$. Under this assumption, even in causal learning, the \rev{cause realizations} change our belief about the ICM parameter $\psi$; namely, we believe it to be similar to $\mu$ estimated from $\D_x$. \rev{As we will show below, any information that leads to updating our belief about the ICM parameter  $\psi$ did not come from the data, but was already incorporated in the joint prior.} For \rev{a more detailed analysis and an illustration of this setting, we refer to Section~\ref{sec:experiments:unsupervised} and Fig.~\ref{fig:unsupervised} below.}
\vspace{1em}

In the remainder of this work we first show in Section~\ref{sec:factorized} that a factorized prior $p(\theta,\psi)=p(\theta)p(\psi)$ results in a  factorized posterior $p(\theta,\psi|\D,\D_x)$, suggesting that factorized priors are an adequate choice for the ICM setting. In Section~\ref{sec:causal} we then show that, regardless of the prior distribution, cause realizations cannot help estimating $\psi$ \emph{beyond what is estimable from an improved estimate of $\theta$}, consolidating the counter-intuitivity of the example with existing theory. 

\section{Causal Semi-Supervised Learning with Factorized Priors}
\label{sec:factorized}

We start our analysis with a factorized prior, i.e., with $p(\theta,\psi)=p(\theta)p(\psi)$. In this setting, it can be shown that the posterior distribution factorizes as well, and that the cause realizations are only effective in the posterior distribution of the cause parameter $\theta$. To see this, note that the posterior distribution $p(\theta,\psi|\D,\D_x)$ is given as
\begin{align}
 p(\theta,\psi|\D,\D_x) &= \frac{p(\theta)p(\psi)\pi(\D,\D_x|\theta,\psi)}{p(\D,\D_x)}\notag\\
 &= \frac{p(\theta)p(\psi)\pi(\Dx|\theta)\pi(\Dy|\Dx,\psi)\pi(\D_x|\theta)}{p(\D,\D_x)} \label{eq:joint_posterior}
\end{align}
where in the second line we made use of~\eqref{eq:pi}.
We next marginalize $p(\D,\D_x,\theta,\psi)$ over $\theta$ and $\psi$ to obtain the denominator:
\begin{align}
 &p(\D,\D_x) \notag\\
 &= \int_\theta \int_\psi p(\theta)p(\psi)\pi(\D,\D_x|\theta,\psi) \mathrm{d}\psi \mathrm{d}\theta\notag\\
 &= \int_\theta \int_\psi p(\theta)p(\psi)\pi(\Dx|\theta)\pi(\Dy|\Dx,\psi)\pi(\D_x|\theta) \mathrm{d}\psi \mathrm{d}\theta\notag\\
  &= \int_\theta \int_\psi p(\psi)\pi(\Dy|\Dx,\psi)\mathrm{d}\psi p(\theta)\pi(\Dx|\theta)\pi(\D_x|\theta)\mathrm{d}\theta \notag\\
 &\stackrel{(a)}{=} \int_\theta \underbrace{\int_\psi p(\psi|\Dx)\pi(\Dy|\Dx,\psi)\mathrm{d}\psi}_{=:p(\Dy|\Dx)} p(\theta)\pi(\Dx|\theta)\pi(\D_x|\theta)\mathrm{d}\theta\notag\\
 &= p(\Dy|\Dx) \int_\theta p(\theta)\pi(\Dx|\theta)\pi(\D_x|\theta)\mathrm{d}\theta\notag \\
 &= p(\Dy|\Dx) p(\Dx,\D_x) \label{eq:marginal}
\end{align}
where in $(a)$ we made use of the fact that 
\begin{multline*}
       p(\psi|\Dx)= \frac{p(\Dx,\psi)}{p(\Dx)}=\frac{p(\psi)p(\Dx|\psi)}{p(\Dx)}\\=\frac{p(\psi)p(\Dx)}{p(\Dx)}=    p(\psi)
\end{multline*}
since $\Dx$ does not depend on $\psi$. Using~\eqref{eq:marginal} in~\eqref{eq:joint_posterior} above yields
\begin{align*}
 p(\theta,\psi|\D,\D_x) &= \frac{p(\theta)p(\psi)\pi(\Dx|\theta)\pi(\Dy|\Dx,\psi)\pi(\D_x|\theta)}{p(\Dy|\Dx) p(\Dx,\D_x)}\\
 &=\frac{p(\theta)\pi(\Dx|\theta)\pi(\D_x|\theta)}{p(\Dx,\D_x)}\cdot\frac{p(\psi)\pi(\Dy|\Dx,\psi)}{p(\Dy|\Dx)}\\
 &= p(\theta|\Dx,\D_x) p(\psi|\Dx,\Dy)\\
 &=p(\theta|\Dx,\D) p(\psi|\D).
\end{align*}

As it can be seen, only fully labeled data $\D$ affects the posterior of the mechanism parameter $\psi$, while both labeled data and \rev{cause realizations} change our belief about the cause parameter $\theta$.

\section{Causal Semi-Supervised Learning with Arbitrary Priors}
\label{sec:causal}

We next investigate how, under a general prior distribution $p(\theta,\psi)$, the posterior distribution $p(\theta,\psi|\D)$ of the cause and ICM parameters changes by including cause realizations. In other words, we investigate the difference between $p(\theta,\psi|\D)$ and $p(\theta,\psi|\D,\D_x)$. We apply the product rule to get
\begin{subequations}\label{eq:product}
    \begin{align}
        p(\theta,\psi|\D) &= p(\theta|\D) p(\psi|\D,\theta)\\
        p(\theta,\psi|\D,\D_x) &= p(\theta|\D,\D_x) p(\psi|\D,\D_x,\theta).
    \end{align}
\end{subequations}
It is obvious that cause realizations will help in estimating the parameter $\theta$ of the cause, i.e., $p(\theta|\D, \D_x)$ will be different from $p(\theta|\D)$. We next show that the second factors on the right-hand sides of~\eqref{eq:product} are equal. Indeed,
\begin{align*}
    p(\psi|\D,\D_x,\theta) &= \frac{p(\psi,\theta,\D,\D_x)}{p(\D,\D_x,\theta)}\\
    &= \frac{p(\psi,\theta)\pi(\D,\D_x|\theta,\psi)}{p(\D,\D_x,\theta)}\\
    &\stackrel{(a)}{=}  \frac{p(\psi,\theta)\pi(\D|\theta,\psi)\pi(\D_x|\theta)}{p(\D,\D_x,\theta)}\\
    &\stackrel{(b)}{=}  \frac{p(\psi,\theta)\pi(\D|\theta,\psi)\pi(\D_x|\theta)}{\pi(\D_x|\theta)p(\D,\theta)}\\
    &= \frac{p(\psi,\theta)\pi(\D|\theta,\psi)}{p(\D,\theta)} = p(\psi|\D,\theta)
\end{align*}
where $(a)$ follows from~\eqref{eq:pi_joint} and~\eqref{eq:pi_dx} and where in $(b)$ we made use of the fact that marginalizing $p(\D,\D_x,\theta,\psi)$ over $\psi$ yields
\begin{multline}
    p(\D,\D_x,\theta) = \int p(\D,\D_x,\theta,\psi) \mathrm{d}\psi \\= \int p(\theta,\psi)\pi(\D|\theta,\psi)\pi(\D_x|\theta) \mathrm{d}\psi\\
    =\pi(\D_x|\theta) \int p(\theta,\psi)\pi(\D|\theta,\psi) \mathrm{d}\psi 
    \\=: \pi(\D_x|\theta) p(\D,\theta).
\end{multline}

Hence, $p(\psi|\D,\D_x,\theta)=p(\psi|\D,\theta)$, from which we conclude that cause realizations $\D_x$ do not tell us anything about the mechanism parameter $\psi$ \emph{beyond what we can learn from a better estimate of the cause parameter $\theta$}. In other words, $\D_x$ can indeed help us update our belief about $\psi$, since it helps us update our belief about $\theta$ and we (initially) believed that $\psi$ and $\theta$ are not independent. There is, however, no direct effect from observing $\D_x$ on our belief about $\psi$ -- any effect is mediated via the parameter $\theta$. Put differently, all the information that makes the marginal posterior $p(\psi|\D,\D_x)$ different from the marginal posterior $p(\psi|\D)$ is already included in the prior $p(\theta,\psi)$.

\begin{figure*}
    \centering
    \includegraphics[width=0.75\textwidth]{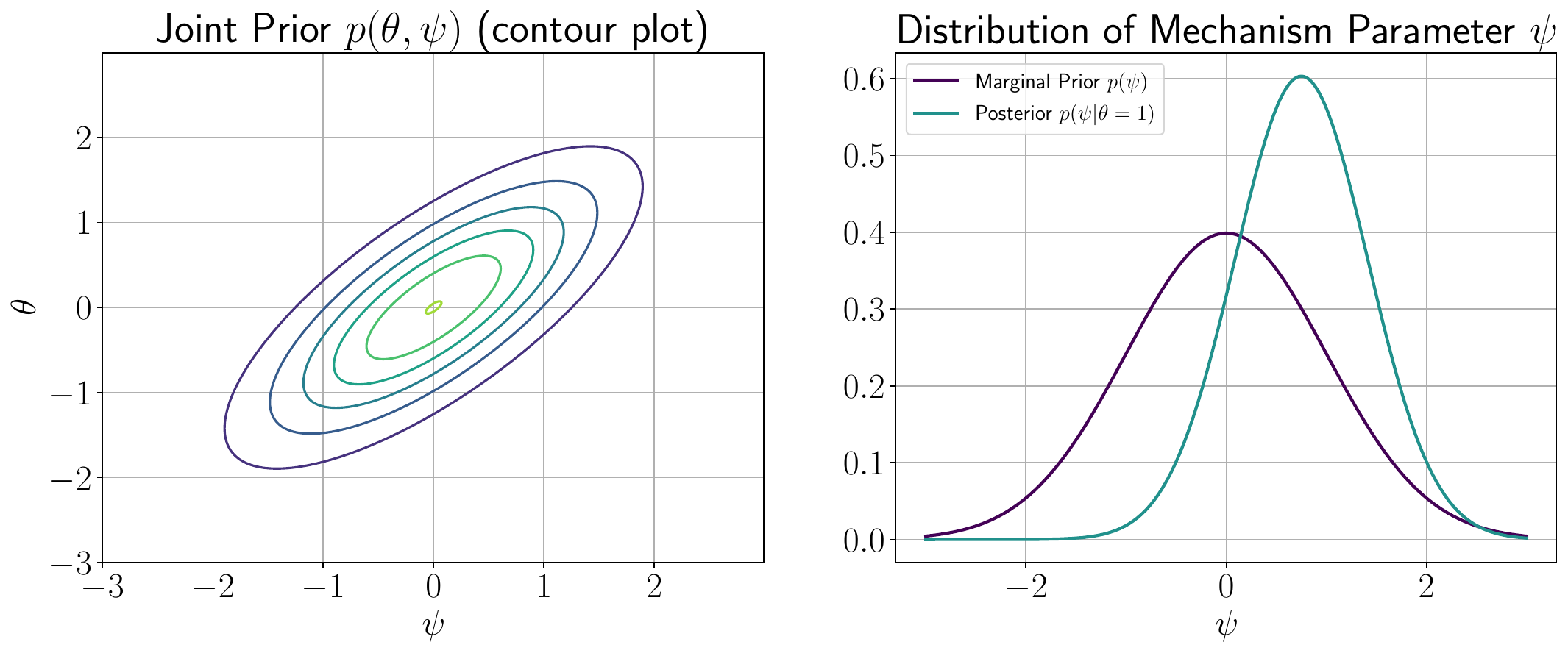}
    \caption{Unsupervised causal learning with \rev{infinitely many cause realizations} ($N=0$ and $M\to\infty$). (Left) The level sets of the prior $p(\theta,\psi)$ are illustrated as a contour plot for $\rho=0.75$. (Right) The prior and posterior distributions of the mechanism parameter $\psi$. Note that the posterior distribution is obtained by evaluating the joint prior at the learned value $\theta=1$.}
    \label{fig:unsupervised}
\end{figure*}

\section{Experiments}
\label{sec:experiments}
We illustrate our findings at the hand of several synthetic examples.\footnote{Code for our experiments can be accessed at \texttt{\url{https://github.com/KNOWSKITE-X/BayesianCausalLearning}}} Specifically, we investigate unsupervised, fully supervised, and semi-supervised settings where our datasets consist of only cause realizations, paired cause and effect realizations, and mixtures thereof, respectively. We conduct these experiments to build intuition about the influence of a correlated prior. More specifically, we show that such a correlated prior not only leads to counterintuitive results as in the Example in Section~\ref{sec:setup}, but that it also slows down learning in fully and semi-supervised settings.

Similar to the Example in Section~\ref{sec:setup}, we consider an additive model $y=x+\eta$. We assume that $x$ and $\eta$ are drawn independently   from Gaussian distributions, with mean $\theta$ and variance 3 and mean $\psi$ and variance 1, respectively. In other words, given the cause and mechanism parameters, the cause and noise realizations are drawn from a Gaussian likelihood $\pi(x,\eta|\theta,\psi)=\mathcal{N}(x,\eta;[\theta,\psi],\Sigma)$ with
\begin{equation}
    \Sigma=\left[ \begin{array}{cc}
        3 & 0 \\
        0 & 1
    \end{array}\right].
\end{equation}
Causal learning of the mechanism $x\to y$ thus requires learning the mean $\psi$ of the Gaussian noise $\eta$. Thanks to the linear model $y=x+\eta$, the labeled dataset $\D$ can be transformed into a dataset $\D'=\{(x_i,\eta_i)\}$ of cause and noise realizations that we will use for the rest of the analysis. Our prior distribution $p(\theta,\psi)$ is Gaussian with zero mean vector $\mu_0=[0,0]$ and covariance matrix
\begin{equation}
    \Sigma_0 =\left[ \begin{array}{cc}
        1 & \rho \\
        \rho & 1
    \end{array}\right]
\end{equation}
where the correlation coefficient $\rho$ represents the strength of dependency between the cause and mechanism parameters that is assumed a priori.

\subsection{Unsupervised Learning}
\label{sec:experiments:unsupervised}
We start with a completely unsupervised setting \rev{that puts the intuition provided in the Example in Section~\ref{sec:setup} on a solid mathematical basis.} In this setting we assume $\D=\D'=\emptyset$ and to have access to \rev{infinitely many cause realizations}, i.e., $M\to\infty$. Thus, under mild assumptions, the posterior $p(\theta|\D_x)$ of the cause parameter converges to a point mass at the true cause parameter $\theta^\bullet$. The posterior for the mechanism parameter is then obtained by evaluating the conditional distribution $p(\psi|\theta)$ obtained from the prior at $\theta^\bullet$. In line with the results in Section~\ref{sec:causal} we therefore have that $p(\psi|\D_x,\theta)=p(\psi|\theta^\bullet)$. 

Fig.~\ref{fig:unsupervised} illustrates this setting for $\theta^\bullet=1$ and a correlation coefficient of $\rho=0.75$. The level sets of the prior are shown as contour lines on the left-hand side, while the prior and posterior distributions of the mechanism parameter $\psi$ are shown on the right-hand side. As it can be seen, the posterior distribution differs substantially from the prior distribution --- \rev{despite the fact that learning relied only on cause realizations. While this appears to be in conflict with the fact that cause realizations are not useful for learning the mechanism, note that here -- as in the Example in Section~\ref{sec:setup} -- any change in belief about the mechanism parameter is simply due to the assumed dependence in the joint prior: The prior distribution of the mechanism parameter is obtained by marginalization, while the posterior distribution is obtained by evaluating the joint prior at $\theta=\theta^\bullet=1$. Hence, any information that leads to updating our belief about the mechanism parameter did not come from the data, but was already incorporated in the joint prior.}

\subsection{Fully Supervised Learning}
\label{sec:experiments:supervised}
As a second setting, we investigate fully supervised learning, i.e., $M=0$ and $\D_x=\emptyset$, but where we have access to a labeled dataset $\D'=\D'_N$ of size $N$. With the joint Gaussian prior $p(\theta,\psi)=\mathcal{N}(\theta,\psi;\mu_0,\Sigma_0)$ parameterized by $\rho$ and the Gaussian likelihood, we obtain a jointly Gaussian posterior~\cite[Sec.~7]{Murphy_GaussianBayes}
\begin{subequations}\label{eq:Gauss_posterior}
  \begin{equation}
    p(\theta,\psi|\D'_N) = \mathcal{N}(\theta,\psi;\mu_N,\Sigma_N)
\end{equation}
where
\begin{align}
    \bar{x} &= \frac{1}{N} \sum_{i=1}^N x_i\\
    \bar{\eta} &= \frac{1}{N} \sum_{i=1}^N \eta_i\\
    \Sigma_N &= \left(\Sigma_0^{-1} + N\Sigma^{-1} \right)^{-1}\\
    \mu_N &= \Sigma_N \left(N \Sigma^{-1} [\bar{x}, \bar{\eta}]^T + \Sigma_0^{-1}\mu_0 \right).
\end{align}  
\end{subequations}

\rev{We conducted the following experiment. For a concrete setting of $\rho$ and $N$, we first draw the true parameters $\mu^\bullet=[\theta^\bullet,\psi^\bullet]$ from the product of marginal prior distributions $p(\theta)p(\psi)$, thus ensuring that the data is generated by an ICM. We then draw $N$ samples of $(x,\eta)$ from the likelihood $\pi(x,\eta|\theta^\bullet,\psi^\bullet)=\mathcal{N}(x,\eta;[\theta^\bullet,\psi^\bullet],\Sigma)$ to populate our dataset $\D'_N$ and use these to update the posterior~\eqref{eq:Gauss_posterior}. We finally evaluate the log-likelihood of the true mechanism parameter under this posterior, i.e., we evaluate $\log p(\psi^\bullet|\D'_N)$. To account for randomness, we draw the true parameters 10,000 times and average the log-likelihood under the posterior.}

\begin{figure}
    \centering
    \includegraphics[width=\linewidth]{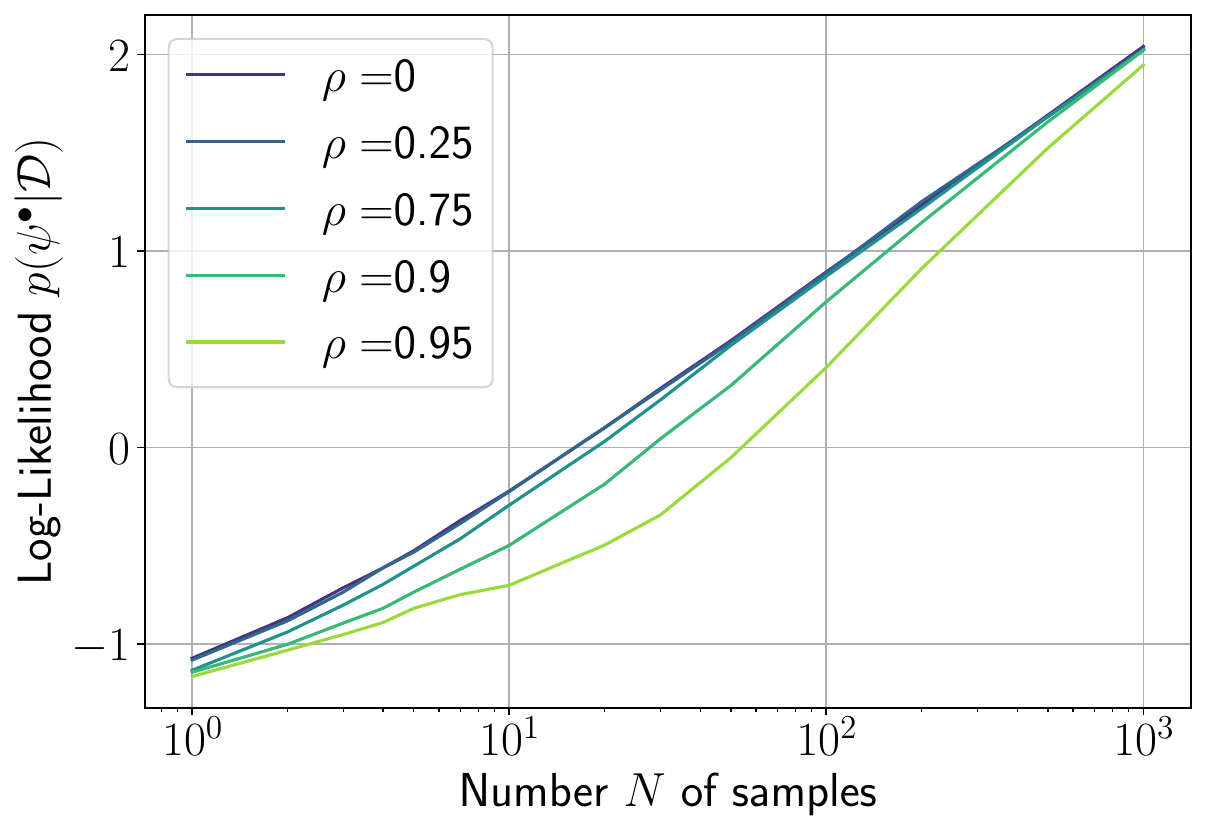}
        \centering
         \includegraphics[width=\linewidth]{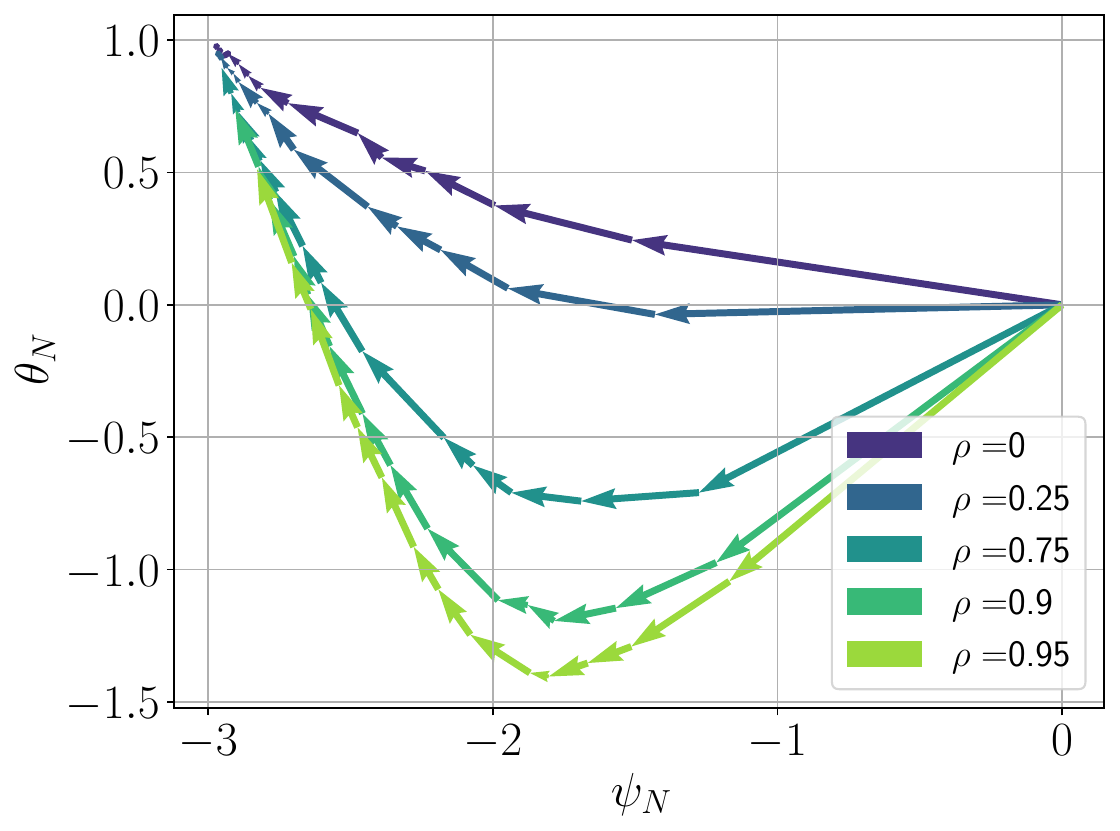}
    \caption{\rev{Supervised causal learning ($M=0$) with randomly chosen cause and effect parameters. (Top) We display the log-likelihood $\log p(\psi^\bullet|\D'_N)$ of the true mechanism parameter as a function of the dataset size $N$, averaged over 10,000 random experiments. The log-likelihood increases with $N$, but slower if the correlation coefficient $\rho$ in the prior is larger. (Bottom) Average trajectories of the posterior means $[\theta_N,\psi_N]$ as a function of $N$. As it can be seen, for a strongly correlated prior, the posterior means take a longer route to reach the true parameters $[\theta^\bullet,\psi^\bullet]=[1,-3]$.}}
    \label{fig:supervised_random}
\end{figure}

The results are shown in Fig.~\ref{fig:supervised_random}. As it can be seen, a strong dependency in the prior (i.e., a large $\rho$) substantially slows down learning in the sense that the \rev{log-likelihood increases much slower} than for a factorized prior ($\rho=0$). \rev{To provide an intuition for this phenomenon, we also plot trajectories of the posterior means $[\theta_N,\psi_N]$ as a function of $N$. We obtained these trajectories by setting the true parameters to $\theta^\bullet=1$ and $\psi^\bullet=-3$, updating the posterior for 1,000 random draws of $(x,\eta)$, and averaging the resulting posterior means $[\theta_N,\psi_N]$. As the plot shows, for large values of $\rho$, the trajectory takes a ``detour'' caused by the fact that the cause and mechanism parameters are pulled in the same direction by the strong prior correlation (in this case, both are decreasing from the respective prior means $\theta_0=0$ and $\psi_0=0$). This detour is particularly strong in the direction of $\theta$, since the} likelihood of the cause parameter has a larger variance, hence benefits less from a given number $N$ of realizations than the mechanism parameter does. In causal learning, such a situation is not unlikely: The mechanism $x\to y$ often \emph{varies less} than the cause, and is in many cases of relevance even deterministic (e.g., in surrogate modeling for deterministic simulations).

\subsection{Semi-Supervised Learning}
\rev{Based on the observations that a strong correlation in the prior slows down fully supervised learning, it is reasonable to assume that this effect is also present semi-supervised settings. Specifically, we believe that for such a correlated prior, additional cause realizations $M>0$ are detrimental in the sense that, for the same size $N$ of the labeled dataset $\D$, the posterior $p(\psi|\D)$ will be strictly more accurate than the posterior $p(\psi|\D,\D_x)$.}

\begin{figure}
    \centering
    \includegraphics[width=\linewidth]{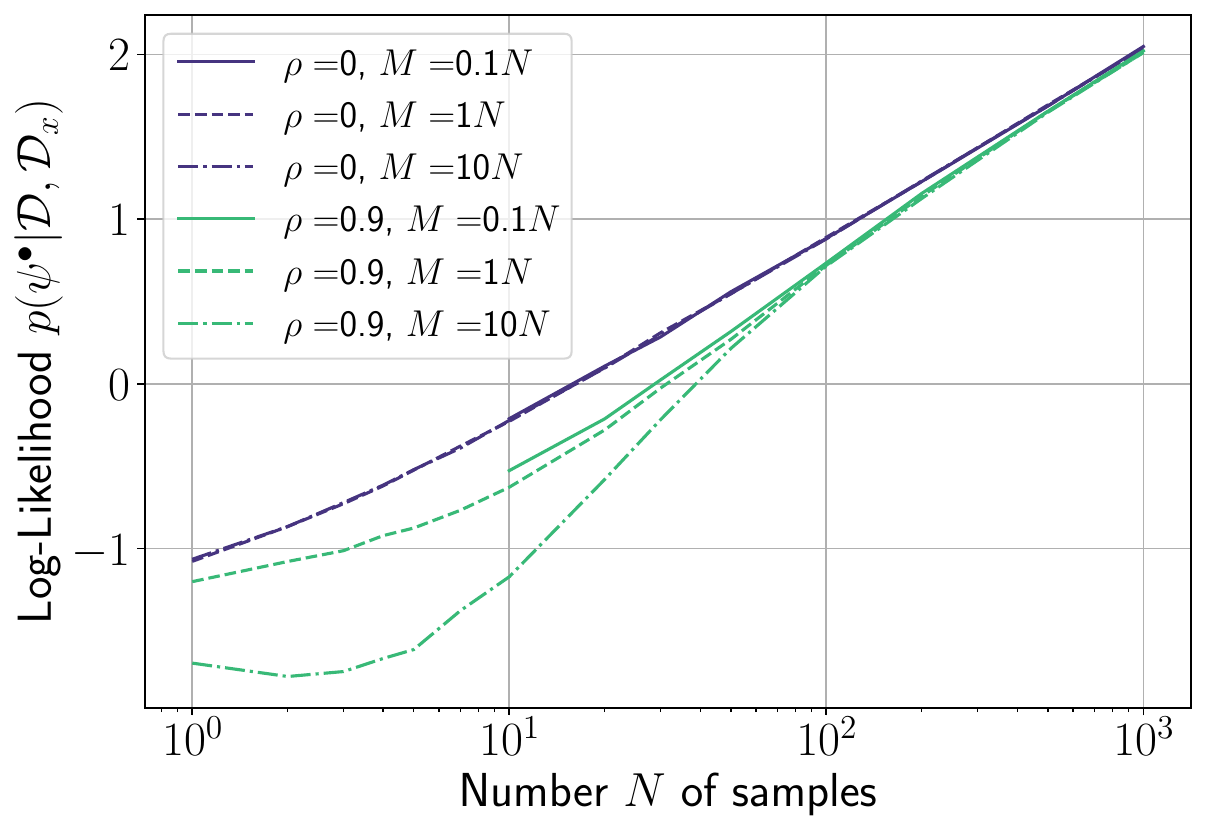}
    \caption{\rev{Semi-supervised causal learning with randomly chosen cause and mechanism parameters. We display the log-likelihood $\log p(\psi^\bullet|\D'_N,\D'_{x,M})$ of the true mechanism parameter as a function of the supervised dataset size $N$ and for different fractions of unsupervised dataset sizes $M$, averaged over 10000 random experiments. Providing additional cause realizations slows down causal learning if the prior is correlated.}}
    \label{fig:semi-supervised}
\end{figure}

\rev{We adhere to the same setting as in Section~\ref{sec:experiments:supervised}. To incorporate a dataset $\D'_{x}=\D'_{x,M}$ of $M$ cause realizations, we adapt the computation of the posterior $p(\theta,\psi|\D'_{x,M}) = \mathcal{N}(\theta,\psi;\mu_{M},\Sigma_{M})$ as follows: We sample $M$ realizations of $(x,\eta)$ from the Gaussian likelihood $\pi(x,\eta|\theta^\bullet,\psi^\bullet)=\mathcal{N}(x,\eta;[\theta^\bullet,\psi^\bullet],\Sigma)$ and compute
\begin{subequations}\label{eq:Gauss_posterior_semi}
\begin{align}
    \bar{x}_M &= \frac{1}{M} \sum_{i=1}^M x_i\\
    \bar{\eta}_M &= \frac{1}{M} \sum_{i=1}^M \eta_i\\
    \Sigma_M &= \left(\Sigma_0^{-1} + M\Sigma' \right)^{-1}\\
    \mu_M &= \Sigma_M \left(M \Sigma' [\bar{x}_M, \bar{\eta}_M]^T + \Sigma_0^{-1}\mu_0 \right)
\end{align}  
with
\begin{equation}
    \Sigma' = \left[ \begin{array}{cc}
        1/3 & 0 \\
        0 & 0
    \end{array}\right],
\end{equation}
\end{subequations}
thus ignoring information from $\bar{\eta}_M$. We then simply update this posterior using a fully supervised dataset $\D'_N$ according to~\eqref{eq:Gauss_posterior}, with $\mu_0$ and $\Sigma_0$ in~\eqref{eq:Gauss_posterior} set to  $\mu_M$ and $\Sigma_M$ , respectively.}

\rev{In our experiments we selected the unlabeled dataset size, i.e., the number $M$ of cause realizations as a fraction or a multiple of the size $N$ of the fully labeled dataset $\D'_N$. While $M=0.1N$ thus corresponds to strong supervision, $M=10N$ corresponds to typical ranges seen in semi-supervised learning.}

\rev{As the results in Fig.~\ref{fig:semi-supervised} show, for an uncorrelated prior the inclusion of cause realizations has no influence on the likelihood of the mechanism parameter under the posterior, as expected. If the prior is correlated, however, we see that not only learning is slowed down (as in Fig.~\ref{fig:supervised_random}), but that larger numbers $M$ of cause realizations slow down learning \emph{more} than smaller numbers. This confirms out hypothesis that for a factorized prior the inclusion of cause realizations is detrimental to learning.}

\section{Discussion}
\label{sec:discussion}

The idea behind an ICM is that it operates on cause realizations independently of their distribution. If one intervenes on the cause (e.g., changing the parameter $\theta$), then the mechanism is not affected and still operates according to its parameterization $\psi$. For example, changing (mildly) the recording setup will change the distribution of recorded audio signals (the cause parameter $\theta$ changes), but not the way how transcripts are produced from the recorded speech (the mechanism parameter $\psi$ does not change). From this interventional perspective, a factorized joint prior for $(\theta,\psi)$ seems reasonable: Even perfect knowledge of the cause parameter $\theta$ (e.g., due to a specific intervention) should not change our prior knowledge about the mechanism we intend to learn. Similarly, even after observing paired cause and effect realizations $\D$, we would not expect that an intervention on the cause substantially changes our belief about the mechanism parameter $\psi$. Hence, we would expect that, in an ICM setting and if learning was successful, the posterior distribution of $(\theta,\psi)$ remains factorized. This, together with our results in Sections~\ref{sec:factorized} and~\ref{sec:causal}, suggests that a factorized prior for $(\theta,\psi)$ is an appropriate choice if one can assume that the mechanism is independent from the cause. \rev{We believe that this insight is particularly relevant in Bayesian deep learning~\cite{Fortuin_2022}, where distributions over (high-dimensional) parameter vectors $(\theta,\psi)$ are often modeled in latent space. In such a case, even if the priors in latent space factorize, special architectures or learning approaches may be necessary to ensure that the corresponding priors (and hence posteriors) also factorize in the high-dimensional spaces of $\theta$ and $\psi$.}

The authors of~\cite{Janzing2010} formulated a definition of ICMs via Kolmogorov complexity, stating that the ICM assumption holds if (in the notation of this work)
\begin{equation}\label{eq:Kolmogorov:discussion}
    I(p(x) : p(y|x)) := K(p(x)) + K(p(y|x)) - K(p(x,y)) \stackrel{+}{=} 0
\end{equation}
where $I(\cdot:\cdot)$ denotes algorithmic mutual information. Assuming that a Turing machine can efficiently transform the description of the cause and mechanism distributions into the parameters that describe them,~\eqref{eq:Kolmogorov:discussion} can be rewritten as
\begin{equation}
    I(p(x) : p(y|x)) \stackrel{+}{=} I(\theta : \psi).
\end{equation}
With~\cite[Th.~2]{Janzing2010} (and ignoring the complexity of evaluating the posterior $p(\theta,\psi|\D,\D_x)$) we obtain that
\begin{equation}
    I(p(x) : p(y|x)) \approx I(\theta;\psi)
\end{equation}
where $I(\cdot;\cdot)$ is the \emph{statistical} mutual information, determined by the distribution from which the parameters $\theta$ and $\psi$ are drawn -- i.e., the posterior $p(\theta,\psi|\D,\D_x)$. Choosing a factorized prior ensures that also this posterior factorizes (cf.~Section~\ref{sec:factorized}), in turn guaranteeing that $I(\theta;\psi|\D,\D_x)=0$. A factorized prior thus also ensures that the algorithmic mutual information between the learned cause and mechanism distributions remains small. This factorization further resonates with the concept of parameter independence in Bayesian inference studied by Heckerman et al. There, however, factorization is not only a consequence of a factorized prior, but also requires fully labeled data, since inference is performed over multiple competing hypothesis about the data generating process (i.e., in the context of this work, about the structural causal model). Here, in contrast, factorization is a result of assuming a factorized prior together with a particular data generating process (namely, an ICM). Studying the interconnection between these independent, but apparently related results is within the scope of future work.

A few words about practical aspects may be in order. While our results confirmed that cause realizations cannot help learning the mechanism, there are considerations that may justify the use of \rev{cause realizations} even in causal learning settings. On the one hand, it is acknowledged that \rev{cause realizations} can help reducing losses or risks used in learning~\cite[Sec.~5.1.2]{Peters_Book}. Indeed, losses are often formulated as averages over the distributions of $x$. In the causal learning setting, having a better estimate of the cause distribution thus allows to learn a model for the mechanism that is better \emph{on average}. On the other hand, in many contemporary problems of practical relevance, the true posterior $p(\theta,\psi|\D,\D_x)$ or predictive posterior $p(y|x,\D,\D_x)$ are intractable, requiring carefully parameterized families of distributions. In some settings, especially with high-dimensional causes, the predictive posterior is parameterized as a learned feature extractor and a task-specific classifier or regressor (as in \rev{natural language processing and} automatic speech recognition, for example). If the feature extractor is obtained via representation learning, then \rev{cause realizations} could enable learning better representations, which could subsequently improve the accuracy of the overall predictive posterior. In other words, even if the true posterior is not affected by \rev{cause realizations}, they may help us finding a model that is closer to the true posterior\rev{; evidence is provided by, e.g.,~\cite[Table 4 \& 5]{Jin_2021} that shows small improvements due to semi-supervised learning even in causal learning settings.} Future work shall investigate this line of argumentation and analyze contemporary semi-supervised learning problems in both causal and anti-causal/confounded settings (similar to~\cite[Fig.~5.2]{Peters_Book}).

\section*{Acknowledgments}
The work was funded by the European Union’s Horizon Europe research and innovation programme within the Knowskite-X project, under grant agreement No. 101091534, and by the Austrian Science Fund, under grant agreement P-32700-NB. Know Center Research GmbH is a COMET center within COMET – Competence Centers for Excellent Technologies. This program is funded by the Austrian Federal Ministries for Climate Policy, Environment, Energy, Mobility, Innovation and Technology (BMK) and for Labor and Economy (BMAW), represented by Österreichische Forschungsförderungsgesellschaft mbH (FFG), Steirische Wirtschaftsförderungsgesellschaft mbH (SFG) and the Province of Styria, Vienna Business Agency and Standortagentur Tirol.

\bibliographystyle{IEEEtran}
\bibliography{references}

\end{document}